\pdfoutput=1

\documentclass[11pt]{article}

\usepackage[preprint]{acl}

\usepackage{times}
\usepackage{latexsym}
\usepackage{bbm, amsbsy, amsmath, amsthm, amsfonts, amssymb, bm}
\usepackage{rotate, array, color, colordvi, psfrag}
\usepackage{hyperref}
\usepackage{hyperref}
\usepackage{mathrsfs}
\usepackage{amsmath}
\usepackage{setspace}
\usepackage{verbatim}
\usepackage{color}
\usepackage{tabularx}
\usepackage{amssymb}
\usepackage{ulem}
\usepackage{booktabs, longtable}
\usepackage{dcolumn}
\usepackage{comment}
\usepackage{float}
\usepackage{setspace}
\usepackage{latexsym}
\usepackage{algorithm}
\usepackage[noend]{algpseudocode}

\usepackage{color, colortbl}
\usepackage{array}
\usepackage{graphicx}
\usepackage{epsfig}
\usepackage{amsmath, bm}
\usepackage{multirow}
\usepackage{multicol}
\usepackage{times}
\usepackage{wrapfig}
\usepackage{xspace}

\usepackage{color}
\usepackage{comment}
\usepackage{ulem}
\usepackage{graphicx}
\usepackage{soul}
\usepackage{tabularray}
\usepackage{ragged2e}

\usepackage{microtype}
\usepackage{graphicx}
\usepackage{subfigure}
\usepackage{booktabs} 
\usepackage{tabularx}
\usepackage{multirow}
\usepackage{pifont}

\usepackage{hyperref}

\usepackage{amsmath}
\usepackage{amssymb}
\usepackage{mathtools}
\usepackage{amsthm}

\usepackage[capitalize, noabbrev]{cleveref}

\usepackage[T1]{fontenc}

\usepackage[utf8]{inputenc}

\usepackage{microtype}

\usepackage{inconsolata}

\usepackage{graphicx}
\newcommand{\ourmethod}{{\fontfamily{lmtt}\selectfont \textbf{GRNFormer}}\xspace}

\usepackage{xargs} 
\usepackage[colorinlistoftodos, prependcaption, textsize=tiny]{todonotes}
\newcommandx{\unsure}[2][1=]{\todo[linecolor=red,backgroundcolor=red!25,bordercolor=red,#1]{#2}}
\newcommandx{\change}[2][1=]{\todo[linecolor=blue,backgroundcolor=blue!25,bordercolor=blue,#1]{#2}}
\newcommandx{\info}[2][1=]{\todo[linecolor=green,backgroundcolor=green!25,bordercolor=green,#1]{#2}}
\newcommandx{\improvement}[2][1=]{\todo[linecolor=Plum,backgroundcolor=Plum!25,bordercolor=Plum,#1]{#2}}

\title{\ourmethod: A Biologically-Guided Framework for Integrating Gene Regulatory
Networks into RNA Foundation Models}

\author{ \textbf{Mufan Qiu\textsuperscript{1}}, \textbf{Xinyu Hu\textsuperscript{2}},
\textbf{Fengwei Zhan\textsuperscript{2,3}}, \textbf{Sukwon Yun\textsuperscript{1}},
 \textbf{Jie Peng\textsuperscript{1}}, \\ \textbf{Ruichen Zhang\textsuperscript{1}},
\textbf{Bhavya Kailkhura\textsuperscript{4}}, \textbf{Jiekun Yang\textsuperscript{2}},
\textbf{Tianlong Chen\textsuperscript{1}} \\ \\ \textsuperscript{1}University of
North Carolina at Chapel Hill, \textsuperscript{2}Rutgers University,\\ \textsuperscript{3}Barnard
College, \textsuperscript{4}Lawrence
 Livermore National Laboratory\\ \small{ \textbf{Correspondence:} \href{tianlong@cs.unc.edu}{tianlong@cs.unc.edu} }
}

\begin{document}
    \maketitle
    \begin{abstract}
        Foundation models for single-cell RNA sequencing (scRNA-seq) have shown promising
        capabilities in capturing gene expression patterns. However, current approaches
        face critical limitations: they \textit{ignore biological prior
        knowledge} encoded in gene regulatory relationships and \textit{fail to
        leverage multi-omics signals} that could provide complementary
        regulatory insights. In this paper, we propose \textbf{\ourmethod}, a
        new framework that systematically integrates multi-scale \textit{Gene
        Regulatory Networks (GRNs)} inferred from multi-omics data into RNA foundation
        model training. Our framework introduces two key innovations. \underline{First},
        we introduce a pipeline for constructing \textit{hierarchical GRNs} that
        capture regulatory relationships at both \textit{cell-type-specific} and
        \textit{cell-specific} resolutions. \underline{Second}, we design a \textit{structure-aware
        integration framework} that addresses the \textit{information asymmetry} 
        in GRNs through two technical advances: \ding{182} A graph topological
        adapter using multi-head cross-attention to weight regulatory relationships
        dynamically, and \ding{183} a novel \textit{edge perturbation strategy}
        that perturb GRNs with biologically-informed co-expression links
        to augment graph neural network training. Comprehensive experiments have
        been conducted on three representative downstream tasks across multiple
        model architectures to demonstrate the effectiveness of \ourmethod. It achieves
        consistent improvements over state-of-the-art (SOTA) baselines:
        $\mathbf{3.6\%}$ increase in drug response prediction correlation, $\mathbf{9.6\%}$
        improvement in single-cell drug classification AUC, and $\mathbf{1.1\%}$
        average gain in gene perturbation prediction accuracy.
    \end{abstract}

    \section{Introduction}
    \label{sec:intro}

    Recent advances in foundation models (FMs) for single-cell RNA sequencing (scRNA-seq)
    analysis has revolutionized our ability to decipher cellular states and gene
    expression patterns. Models like scGPT~\cite{cui2024scgpt}, Geneformer~\cite{theodoris2023transfer},
    and scFoundation~\cite{hao2024large} demonstrate remarkable capabilities in
    capturing transcriptomic relationships through large-scale pretraining on
    millions of cells. These models achieve state-of-the-art performance in critical
    tasks, including cell type annotation, perturbation prediction, and multi-omic
    integration. Particularly noteworthy is scPaLM~\cite{chen2024pre}, which
    introduces biological pathway-aware representations to address computational
    challenges in transformer-based approaches.

    \begin{figure}[t]
        \centering
        \includegraphics[width=0.9\columnwidth]{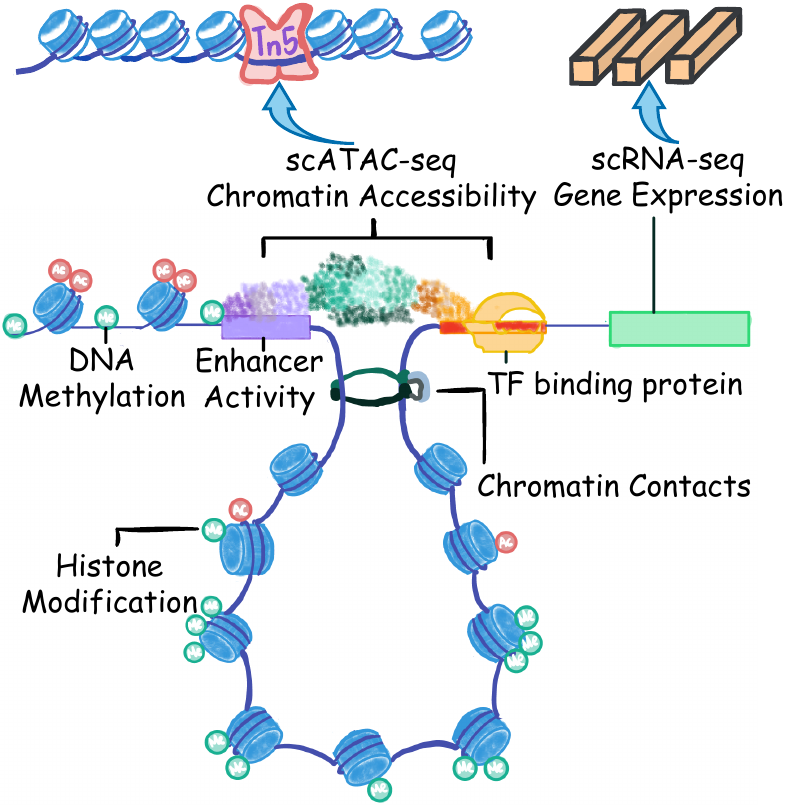}
        \vspace{-3mm}
        \caption{\small{Gene regulatory process in scATAC-seq and scRNA-seq modalities. Image credit to \citet{bonev2024opportunities}.}}
        \label{fig:teaser}
        \vspace{-6mm}
    \end{figure}

    However, despite their successes, current RNA FMs face fundamental limitations
    rooted in their reliance on expression data alone. Three key challenges
    persist in existing approaches. \underline{First}, as shown in Fig.~\ref{fig:teaser},
    while current models learn gene-gene correlations implicitly, they lack
    explicit integration of \textit{regulatory causality} derived from chromatin
    accessibility data -- a crucial determinant of cellular identity~\cite{bravo2023scenic+}.
    \underline{Second}, existing methods struggle to capture the multi-scale nature
    of gene regulation, where relationships operate at both \textit{cell-type-specific} and
        \textit{cell-specific}~\cite{kamimoto2020celloracle}. \underline{Third},
    severe \textit{information asymmetry} plagues regulatory networks: for some
    cell types, transcription factors (TFs) exhibit dense connectivity while $\sim
    {}40\%$ of genes lack reliable regulatory links~\cite{aibar2017scenic, bravo2023scenic+},
    creating a topological imbalance that standard architectures cannot
    effectively handle~\cite{chen2021topology}.
    \begin{figure*}[t]
        \centering
        \includegraphics[width=1\textwidth]{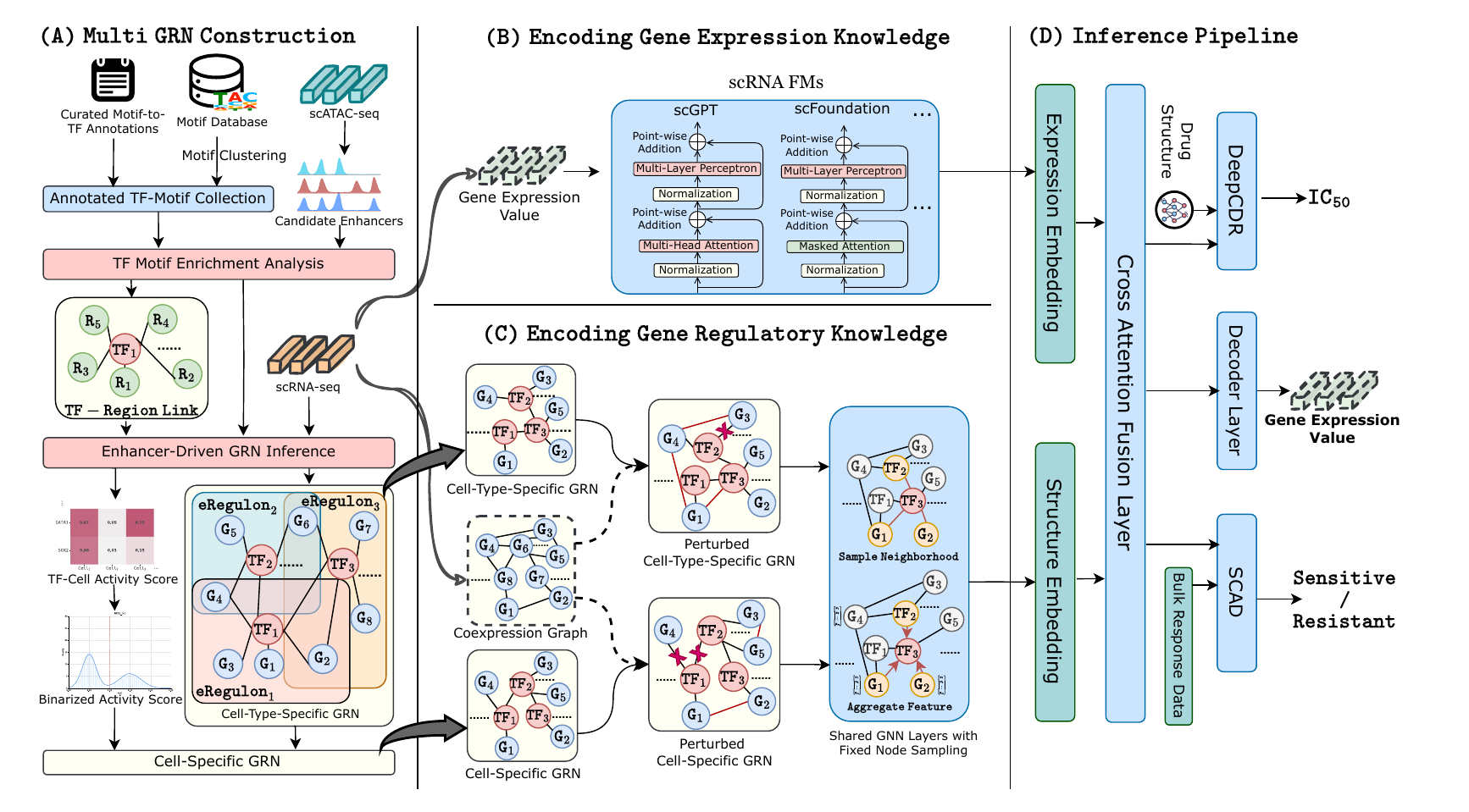}
        \vspace{-9mm}
        \caption{Overview of \ourmethod framework: (A) Multi-scale GRN construction
        from scATAC/scRNA-seq data utilizing additional Motif databases; (B) Our
        framework employs single-cell RNA foundation models (scRNA FMs) to
        encode gene expression profiles into expression embeddings, supporting three
        model architectures as backbones: \textit{scGPT}, \textit{scFoundation},
        and \textit{scPaLM}; (C) The multi-scale GRNs are perturbed using co-expression
        graphs and subsequently processed through GNN modules, with the resulting
        embeddings aggregated via summation to generate the structure embedding;
        (D) The expression embedding and structure embedding obtained from the
        previous two stages are fused through a cross-attention layer. The resulting
        hybrid embedding can be fed into the decoder for pretraining via masked language
        modeling objectives, or directly utilized for diverse downstream tasks.}
        \label{fig:overview}
        \vspace{-3mm}
    \end{figure*}

    To handle these challenges, we present \ourmethod, a novel architecture that
    integrates multi-scale Gene Regulatory Networks (GRNs) into current
    RNA FMs through three key innovations. First, we introduce a systematic pipeline
    for constructing \textit{cell-specific} and \textit{cell-type-specific} GRNs
    capturing \textit{regulatory causality} derived from chromatin accessibility
    data by integrating single-cell ATAC-seq (scATAC-seq) and scRNA-seq data utilizing
    SCENIC+~\cite{bravo2023scenic+}. As shown in Fig.~\ref{fig:overview}A and Appendix~\ref{appendix:eregulon},
    our method leverages chromatin accessibility to identify
    {enhancer-driven regulatory units} (eRegulons) through motif enrichment
    analysis and multi-modal linkage~\cite{bravo2023scenic+}, enabling discovery
    of {context-specific} regulatory relationships across biological scales.

    Then, we introduce a {universal structure-aware integration framework} that
    utilizes the multi-scale gene regulation information and addresses GRNs
    topological challenges through: $i$) an adaptive cross-attention layer that dynamically
    weights regulatory signals based on node centrality and $ii$) a biologically
    informed edge perturbation strategy that supplements sparse connections with
    co-expression relationships as shown in Fig.~\ref{fig:overview}C. This design
    enables effective knowledge transfer from GRNs while mitigating
    \textit{information asymmetry} -- a critical advancement over naive fusion
    approaches such as addition or concatenation.

    Lastly, we establish comprehensive benchmarks across three clinically-relevant
    tasks: gene perturbation prediction, drug response classification, and
    single-cell sensitivity analysis. Our experiments demonstrate that \ourmethod
    achieves consistent improvements over base models (scGPT $+3.6\%$ Pearson
    Correlation Coefficient (PCC) on the drug response prediction task, scFoundation
    $+4.1\%$ Area Under the ROC Curve (AUC) on the single cell drug response classification
    task). Notably, the model reveals interpretable attention patterns aligning
    with known biological regulations. Our key contributions are three folds:
    \begin{itemize}
        \item[\ding{182}] {\textit{Multi-scale GRN Construction Pipeline:}} The first
            systematic framework integrating scATAC-seq and scRNA-seq data to build
            cell-type-specific and single-cell-resolution regulatory networks through
            enhancer-driven eRegulons analysis pipelines.

        \item[\ding{183}] {\textit{Structure-aware Model Architecture:}} An integration
            strategy combining adaptive cross-attention with novel biological guided
            edge perturbation strategy, effectively resolving GRNs topological
            imbalance while maintaining computational efficiency.

        \item[\ding{184}] {\textit{Extensive Biological Validation:}} State-of-the-art
            performance across three therapeutic development tasks, with
            demonstrated improvements in drug response prediction (\textit{e.g.},
            $3.6\%$ of $\mathrm{PCC_{delta}}$ gain against baselines) and single
            cell drug sensitivity classification (\textit{e.g.}, $0.122$ of AUC gain
            against baselines)

    \end{itemize}

    The success of \ourmethod underscores the transformative potential of integrating
    regulatory prior knowledge from different modalities into foundation models.
    Our work establishes a new paradigm for developing biologically grounded AI
    systems in computational genomics, with immediate applications in the discovery
    of drug targets and the improvement of existing gene therapies.

    \section{Related Works}
    \label{sec:rna_fm_related}
    \noindent
    \textbf{Single-cell Data Analysis.} Single-cell RNA sequencing (scRNA-seq)
    has revolutionized genomics by enabling profiling of cell-level gene
    expressions \cite{saliba2014single, kolodziejczyk2015technology}. Providing hints
    on cellular heterogeneity, scRNA-seq transforms how to understand complex
    biological systems such as neural tissues, immune responses, and tumor micro-environments.
    Advances in perturbation sequencing techniques, such as Perturb-seq, have further
    allowed researchers to discover the causal relations between gene perturbations
    and cellular phenotypes by utilizing CRISPR-based editing alongside scRNA-seq
    \cite{dixit2016perturb, adamson2016multiplexed, norman2019exploring}.
    However, integrating information from other omics modalities, such as scATAC-seq
    or spatial omics, remains a significant challenge despite the remarkable progress
    in scRNA-seq technologies~\cite{cui2023scgpt, xiong_scclip_2023}.

    \noindent
    \textbf{Foundation Models in Single-cell Omics.} FMs, first developed for natural
    language processing, have become powerful tools for learning hidden
    embeddings of large-scale biological data. These models, typically
    pretrained on vast datasets, can be fine-tuned for downstream tasks such as classifications
    and translations, offering extensive flexibility and scalability~\cite{bommasani2021opportunities,
    moor2023foundation}. In single-cell biology, foundation models are pre-trained
    on large single-cell datasets, and then applied to downstream tasks like cell
    type annotation, perturbation prediction, and multi-omic integration~\cite{cui2023scgpt,
    theodoris2023transfer}. During the fine-tuning process, model parameters are
    further optimized using task-specific datasets typically of much smaller size
    than the training data, resulting in much lower computational cost\cite{gururangan2020dont,
    qiu2020pretrained}. Additionally, foundation models can recognize various data
    types , such as transcriptomics and epigenomics, providing a more
    generalized view of cell biology~\cite{brown2020language, openai2023gpt4}. 

    Modern scRNA-seq generates gene expression profiles as a cell-by-gene matrix
    $X \in \mathbb{R}^{N \times G}$, where each element $X_{ij}$ represents the
    expression count of gene $j$ in cell $i$. RNA foundation models typically employ
    masked language modeling objectives adapted to transcriptomic data. Given an
    input expression vector $x \in \mathbb{R}^{G}$, these models randomly mask a
    subset of genes $\mathcal{M}\subset \{1,...,G\}$ and optimize reconstruction
    via $\mathcal{L}= \mathbb{E}_{x}\left[\sum_{i \in \mathcal{M}}||f_{\theta}(x^{\text{masked}}
    )_{i}- x_{i}| |^{2}\right]$, where $f_{\theta}$ denotes the foundation model.
    Key architectural variants include: ($1$) \textbf{scGPT}\cite{cui2024scgpt}
    employs generative pretraining with specialized attention masking for non-sequential
    omics data; ($2$) \textbf{scFoundation} \cite{hao2024large} introduces a
    read-depth-aware (RDA) pretraining task using an asymmetric transformer
    architecture. \textbf{scPaLM} \cite{chen2024pre} also tries extending
    current architecture through pathway-aware architectures. Despite these architectural
    explorations, current foundation models remain predominantly focused on
    scRNA-seq data, lacking systematic integration of multi-omics signals such as
    chromatin accessibility profiles from scATAC-seq data.
    \section{Methodology -- \ourmethod}
    \noindent
    \textbf{Overview of \ourmethod.} Our approach addresses the challenge of integrating
    biological prior knowledge of RNA foundation models through a two-stage framework
    as shown in Fig.~\ref{fig:overview}. First, we leverage multi-omics data to
    construct reliable gene regulatory networks (GRNs) at multiple scales - \textit{cell-specific}
    and \textit{cell-type-specific} levels. These networks capture the complex regulatory
    relationships between transcription factors and their target genes. Second, we
    develop a structure-aware integration mechanism that uses cross-attention to
    incorporate GRNs information into RNA foundation model training while handling
    the inherent sparsity and topological imbalance of regulations.

    \subsection{Construction of Multi-scale GRNs}
    \label{sec:grn_construction} Gene regulatory networks (GRNs) from the computational
    blueprint of cellular identity, encoding how transcription factors (TFs) –
    proteins that bind DNA to control gene expression – orchestrate
    transcriptional programs through \textit{cis}-regulatory elements.
    Traditional GRN inference methods face two critical limitations: ($1$) reliance
    on expression correlations alone, missing causal chromatin accessibility
    signals; ($2$) inability to resolve regulatory relationships at both population
    (cell type) and single-cell levels~\cite{aibar2017scenic}. Our framework
    addresses these through multi-modal integration and multi-scale analysis as
    shown in Fig.~\ref{fig:overview} A. We first begin with cell-type-specific
    GRNs generation, which we mainly followed SCENIC+~\cite{bravo2023scenic+}
    pipeline and the details can be found at Appendix~\ref{appendix:eregulon}.

    \noindent
    \textbf{Single-cell GRNs via Activity Thresholding.} To resolve regulatory
    heterogeneity within cell types, we quantify eRegulon activity at single-cell
    resolution using AUCell~\cite{aibar2017scenic}. This algorithm calculates an
    \textit{Area Under the recovery Curve} (AUC) score by ranking genes or regions
    and measuring target set enrichment. Critically, the AUC distribution across
    cells reveals fundamental biological patterns: ($1$) \textbf{Bimodal
    distributions} indicate two distinct cell subpopulations (active/inactive),
    while ($2$) \textbf{Skewed Gaussian distributions} reflect graded activation
    across a continuum~\cite{van2020scalable}. 
    We model these patterns using a two-component Gaussian mixture:
    \begin{equation}
        p(x) = \pi_{1}\mathcal{N}(x|\mu_{1},\sigma_{1}^{2}) + \pi_{2}\mathcal{N}(
        x|\mu_{2},\sigma_{2}^{2})
    \end{equation}
    where $\pi_{i}$ are mixing coefficients. For bimodal cases, the threshold is
    set at the Gaussians' intersection, cleanly separating active and inactive
    cells. For skewed distributions with a single dominant component, we label
    cells in the right tail ($\mu + 2\sigma$) as active, capturing cells with exceptionally
    strong regulon activity. This biologically-grounded thresholding ensures
    each cell's GRN comprises only context-relevant regulatory interactions. Examples
    of the activity distribution of transcription factors and the corresponding thresholds
    in our pre-training data are illustrated in Appendix \ref{appendix:tf_dist}.

    \noindent
    \textbf{Cross-modality Integration.} Recognizing that most downstream tasks
    involve single-modality scRNA-seq datasets, we enable GRN integration through
    reference mapping. For single omics downstream datasets, we leverage embeddings
    from pre-trained single multi-omics foundation models (scGPT~\cite{cui2024scgpt},
    scFoundation~\cite{hao2024large}) to map query cells to their nearest
    neighbors in the reference space. This method establishes connections between
    downstream cells and precomputed multi-scale GRNs from paired scATAC-seq and
    scRNA-seq data, ensuring broad applicability across diverse biological
    contexts.

    \subsection{Structure Adapter to Incorporate Gene Regulation}
    \label{sec:structure_adapter} Our structure-aware integration framework
    focuses on addressing three fundamental challenges in incorporating multi-scale
    GRNs: ($1$) \textit{topological imbalance} where TFs dominate connectivity
    while $\sim 40\%$ of genes lack reliable regulations~\cite{aibar2017scenic};
    ($2$) \textit{information asymmetry} between TF-rich and isolated gene
    representations; ($3$) \textit{multi-scale regulatory dynamics} requiring
    simultaneous modeling of cell-type and single-cell contexts~\cite{bravo2023scenic+}.

    \noindent
    \textbf{Architecture Adaptation for Different Backbones.} As shown in Fig.~\ref{fig:overview}B,
    our framework utilizes existing RNA FMs to encode gene expressions and
    demonstrates universal applicability across major variants: ($1$) For \textit{decoder-only}
    models (\textit{i.e.}, scGPT~\cite{cui2024scgpt}), we utilize the embedding before
    the last transformer layer as our expression embedding; ($2$) For \textit{encoder-decoder}
    models (\textit{i.e.}, scFoundation~\cite{hao2024large}, scPaLM~\cite{chen2024pre}),
    we fuse structural embeddings with encoder outputs before feeding them to the
    decoder.

    \noindent
    \textbf{Multi-scale GRN Processing.} Following the pipeline in Section~\ref{sec:grn_construction},
    we process cell-specific and cell-type-specific GRNs using GraphSAGE~\cite{hamilton2017inductive},
    chosen for its ability to handle \textit{degree imbalance} through fixed-size
    neighborhood sampling as shown in Fig.~\ref{fig:overview}C. Traditional GNNs
    that aggregate all neighbors would amplify magnitude differences between
    high-degree TFs (average degree $81.3$ in our data) and remaining genes (average
    degree $1.3$ in our data). For each node $v$ at layer $k$, the aggregation
    follows:

    \begin{equation}
        \begin{aligned}
            h^{k}_{\mathcal{N}(v)} & = \textsc{aggregate}_{k}\left(\{h_{u}^{k-1}, \forall u \in \mathcal{N}(v)\}\right)                \\
            h_{v}^{k}              & = \sigma \left(W^{k}\cdot \textsc{concat}\left(h_{v}^{k-1}, h^{k}_{\mathcal{N}(v)}\right)\right),
        \end{aligned}
    \end{equation}
    where $\mathcal{N}(v)$ denotes a fixed-size uniform sample of neighbors, addressing
    degree imbalance through neighbor sampling as in~\citet{hamilton2017inductive}.
    The final structural embedding $h_{\text{struct}}= h_{\text{cell}}\oplus h_{\text{type}}$
    combines regulation information at both scales through element-wise summation.

    \noindent
    \textbf{Cross-modal Fusion.} Direct concatenation of GRN embeddings ($h_{\text{struct}}$)
    with expression features ($h_{\text{expr}}$) amplifies information asymmetry.
    Instead, our multi-head cross-attention dynamically reweights features. The query-key
    mechanism prioritizes TF-gene interactions with high topological centrality while
    attenuating noise from unconnected genes. This mechanism produces context-aware
    fusion embedding $h_{\text{fusion}}$ that complements expression patterns
    with regulatory constraints.

    \noindent
    \textbf{Edge Perturbation for Topological Balance.} Conventional graph augmentations~\cite{zhao2022graph}
    risk involving biologically meaningless connections. Our \textit{biologically-informed
    perturbation} replaces $\alpha|E|$ edges ($\alpha=0.2$) with co-expression links
    from $G_{\text{co}}$, constructed per cell as:
    \begin{equation}
        G_{\text{co}}= \{(u,v) | x_{u}> 0 \land x_{v}> 0\}, \forall u,v \in \mathcal{G}
    \end{equation}
    where $x$ denotes normalized gene expression, $\mathcal{G}$ denotes the gene
    vocabulary. This perturbation strategy preserves connectivity for genes lacking
    regulatory annotations while maintaining biological plausibility – co-expressed
    genes in the same cell are more likely to share functional relationships~\cite{van2020scalable,
    roohani2022gears}. Compared to random edge perturbation, our approach ensures
    that node embeddings for all non-zero-expressed genes receive sufficient training
    through the sampling of co-expression graph. 

    \subsection{Pretraining and Inference Pipeline}
    The training and inference pipeline of our model is illustrated in Fig.~\ref{fig:overview}D.
    For each backbone architecture, the pretraining objectives and data
    processing pipelines remain consistent with their original implementations, which
    primarily involve variants of masked language modeling tasks.

    We implemented downstream task pipelines based on scGPT and scFoundation frameworks,
    with additional integration of scPaLM.
    Detailed descriptions of these downstream task workflows are provided in the
    \nameref{sec:experiments} Section. \label{sec:inference_pipeline}

    \section{Experiments}
    \label{sec:experiments} We conducted extensive experiments to evaluate
    \ourmethod across three biologically significant tasks: \ding{182} \textbf{Gene
    perturbation prediction} examines the model's ability to capture regulatory mechanisms
    by predicting gene expression changes following gene perturbations. This
    task is particularly relevant for therapeutic development and understanding
    disease mechanisms. \ding{183} \textbf{Drug response prediction} evaluates the
    model's clinical utility by predicting cellular responses to therapeutic compounds.
    The model integrates gene expression profiles with drug structural
    information to predict IC50 values (half-maximal inhibitory concentrations).
    \ding{184} \textbf{Single-cell drug response classification} tests the model's
    ability to transfer knowledge from bulk cell line to single-cell resolution,
    a critical capability for personalized medicine. The task involves predicting
    drug sensitivity for individual cells. Across all these tasks, we will compare
    our approach against SOTA baselines and conduct comprehensive ablation studies
    to evaluate our GRN integration strategy's effectiveness systematically.
    This multi-faceted evaluation framework ensures a thorough assessment of our
    approach's biological accuracy and practical utility.

    \subsection{Implementation Details.}
    \noindent
    \textbf{Pretraining Data.} We pre-trained our model using the Seattle Alzheimer's
    Disease Brain Cell Atlas (SEA-AD) dataset~\cite{hawrylycz2024sea}, which
    provides paired scRNA-seq and scATAC-seq measurements for \textbf{$113, 209$
    cells} from \textbf{$28$ donors}. The scRNA-seq data captures expression profiles
    for \textbf{$18,984$ protein-coding genes}, while the scATAC-seq data provides
    chromatin accessibility information across the genome. Detailed statistics
    about the dataset can be found in Appendix~\ref{appendix:datasets}.

    \noindent
    \textbf{Architectures.} Our framework comprises three core components: A
    \textit{transformer-based RNA foundation model} backbone processing gene expression
    embeddings; A \textit{GraphSAGE encoder}~\cite{hamilton2017inductive} generating
    gene structural embeddings from multi-scale GRNs; and A \textit{cross-attention
    fusion layer} replacing the final transformer layer to integrate structural and
    expression features. The architecture preserves the original backbone
    dimensions (\textit{e.g.}, $768$ hidden units for scFoundation). 

    \noindent
    \textbf{Training Settings.} For 
    scGPT~\cite{cui2024scgpt} and scPaLM~\cite{chen2024pre} backbone, we
    conducted full pretraining on SEA-AD multiome data~\cite{hawrylycz2024sea}. For
    scFoundation~\cite{hao2024large} backbone, we performed continued
    pretraining from their public checkpoint, validating our method's \textit{plug-and-play}
    capability. All models used backbone-specific hyperparameters from original implementations,
    including optimizer type, learning rate, and batch size. Training completed on
    $8\times$A100 GPUs with full reproducibility. Details of the pretraining algorithm
    with multi-scale GRNs can be found in Appendix~\ref{appendix:alg}.

    \noindent
    \textbf{Benchmarks Data.} We established three evaluation paradigms: ($1$) \textit{Gene
    perturbation prediction} using Adamson~\cite{adamson2016multiplexed}($8 7$
    single gene perturbations in protein response pathway), Dixit~\cite{dixit2016perturb}
    (single and combinatorial LPS response gene perturbations), and Norman~\cite{norman2019exploring}
    ($131$ gene pairs and $105$ single genes in K562 cells) datasets; ($2$)
    \textit{Bulk drug response prediction} via CCLE~\cite{barretina2012cancer} ($2
    4$
    drugs, $947$ cell lines) and GDSC~\cite{iorio2016landscape} ($297$ compounds,
    $969$ cell lines); ($3$) \textit{Single-cell drug classification} following scFoundation's~\cite{hao2024large}
    protocol for four commonly cancer targeted therapies (Sorafenib, NVP-TAE684,
    PLX4720, Etoposide). Detailed statistics about these datasets can be found in
    Appendix~\ref{appendix:datasets}.

    \noindent
    \textbf{Baselines.} We established three fundamental baselines: Our implementations
    of scGPT~\cite{cui2024scgpt} and scPaLM~\cite{chen2024pre} pre-trained on SEA-AD
    multiome data, and the officially pre-trained scFoundation~\cite{hao2024large}
    checkpoint. For drug response prediction, we additionally compared against
    DeepCDR~\cite{liu2020deepcdr} as a specialized baseline. For single-cell sensitivity
    classification, we included SCAD~\cite{roohani2022gears} to benchmark cell
    resolution capabilities. All scFoundation results report the maximum performance
    between its original pre-trained version and our continued pretraining variant
    for fair comparison.

    \subsection{Gene Perturbation Prediction}

    Gene perturbation prediction represents a critical task in computational
    biology with direct implications for therapeutic development and disease
    understanding. The task involves predicting genome-wide transcriptional changes
    following genetic interventions, which is essential for understanding gene
    function and identifying potential drug targets. A key challenge in this task
    is capturing the complex, non-linear effects of gene perturbations on
    cellular transcriptional programs.

    \begin{table}[htbp]
        \centering
        \caption{Gene perturbation prediction evaluation.}
        \vspace{-3mm}
        \label{tab:gene_perturb} \resizebox{\linewidth}{!}{\begin{tabular}{lccc|c}
            \toprule Model                              & Adamson          & Dixit            & Norman           & Avg. $\mathrm{PCC_{delta}}$ $\uparrow$    \\
            \midrule scGPT                              & $0.609$          & $0.130$          & $0.405$          & $0.381$\scriptsize{$\pm{0.240}$}          \\
            \rowcolor[gray]{0.90} + \textbf{GRN (ours)} & $\textbf{0.622}$ & $\textbf{0.138}$ & $\textbf{0.418}$ & $\textbf{0.393}$\scriptsize{$\pm{0.243}$} \\
            \midrule scFoundation                       & $0.483$          & $0.239$          & $0.255$          & $0.326$\scriptsize{$\pm{0.137}$}          \\
            \rowcolor[gray]{0.90} + \textbf{GRN (ours)} & $\textbf{0.487}$ & $\textbf{0.241}$ & $\textbf{0.283}$ & $\textbf{0.337}$\scriptsize{$\pm{0.132}$} \\
            \bottomrule
        \end{tabular}}
        \vspace{-3mm}
    \end{table}

    Our evaluation utilized three widely-used benchmark datasets (Adamson ~\cite{adamson2016multiplexed},
    Norman~\cite{norman2019exploring}, and Dixit~\cite{dixit2016perturb}). The input
    comprises unperturbed gene expression profiles and perturbation gene targets,
    while the output comprises predicted post-perturbation expression levels. We
    focused on the Pearson correlation coefficient on differential expression ($\mathrm{PCC_{delta}}$),
    which measures how well the model predicts expression changes directions.

    As shown in Table~\ref{tab:gene_perturb}, \ourmethod achieves consistent improvements
    across all datasets. The GRN-enhanced scGPT variant attains a $1 .1\%$
    average PCC increase ($0.393$ vs. $0.381$ baseline), with particularly
    robust gains on the Norman dataset ($+3.1\%$). We adapted each model's native
    pipeline for gene perturbation prediction, with critical divergence in fine-tuning
    strategies: \textit{scFoundation} employed parameter freezing for most
    layers due to GPU memory constraints, while \textit{scGPT} permitted full
    parameter updates. This architectural distinction likely contributes to scFoundation's
    relatively lower performance, as partial fine-tuning may limit its
    adaptability to perturbation patterns.

    \subsection{Cancer Drug Response Prediction}

    Accurate prediction of cancer drug responses enables personalized treatment strategies
    and accelerates therapeutic development~\cite{barretina2012cancer,iorio2016landscape}.
    We evaluate our model on CCLE and GDSC datasets using IC50 values (half-maximal
    inhibitory concentration) as ground truth. All experiments were repeated four
    times with identical settings except for random seed variations, with means
    and standard deviations calculated. We integrate gene expression profiles with
    drug structural information through DeepCDR-style architecture~\cite{liu2020deepcdr,
    hao2023large}.

    \begin{figure}[t]
        \centering
        \includegraphics[width=0.95\columnwidth]{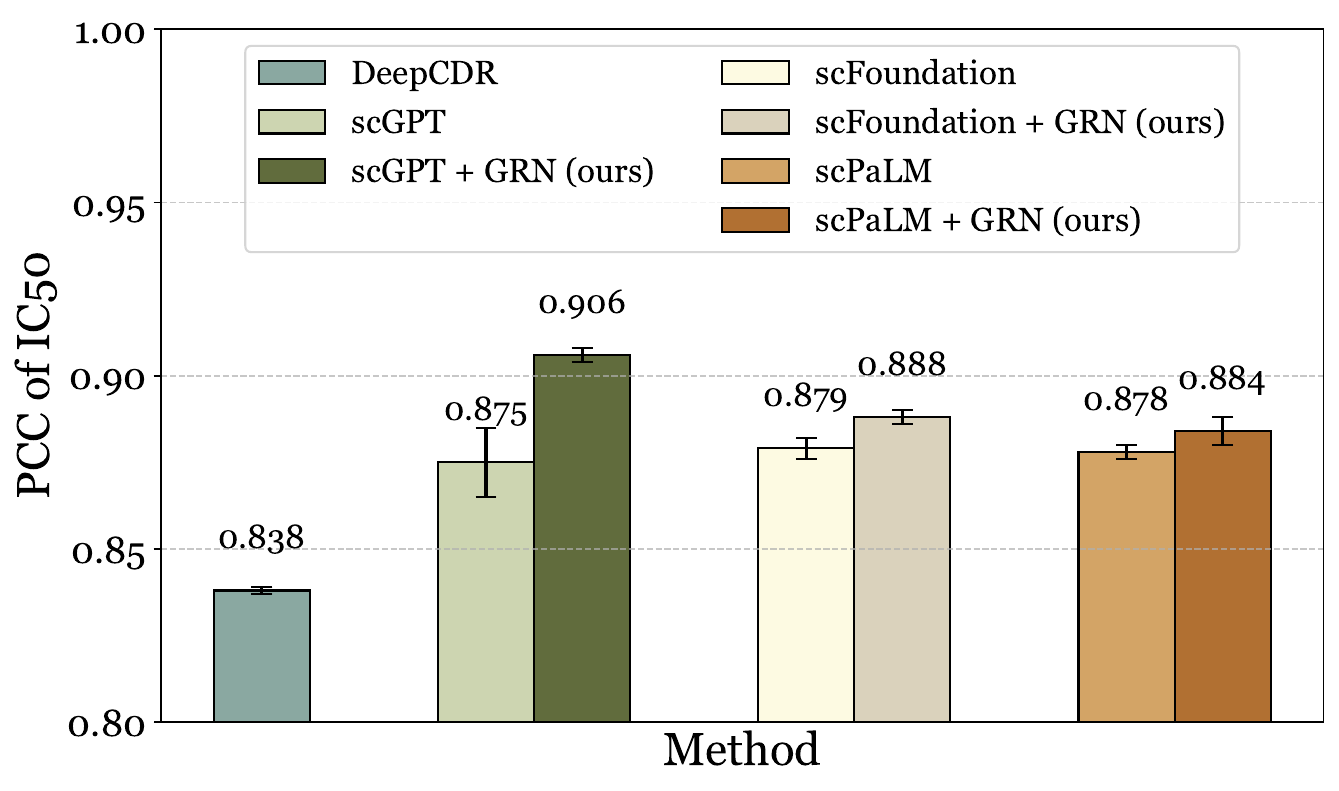}
        \vspace{-4mm}
        \caption{Cancer drug response prediction evaluation.}
        \label{fig:drug_response_num}
        \vspace{-3mm}
    \end{figure}

\begin{figure*}[t]
        \centering
        \includegraphics[width=\textwidth]{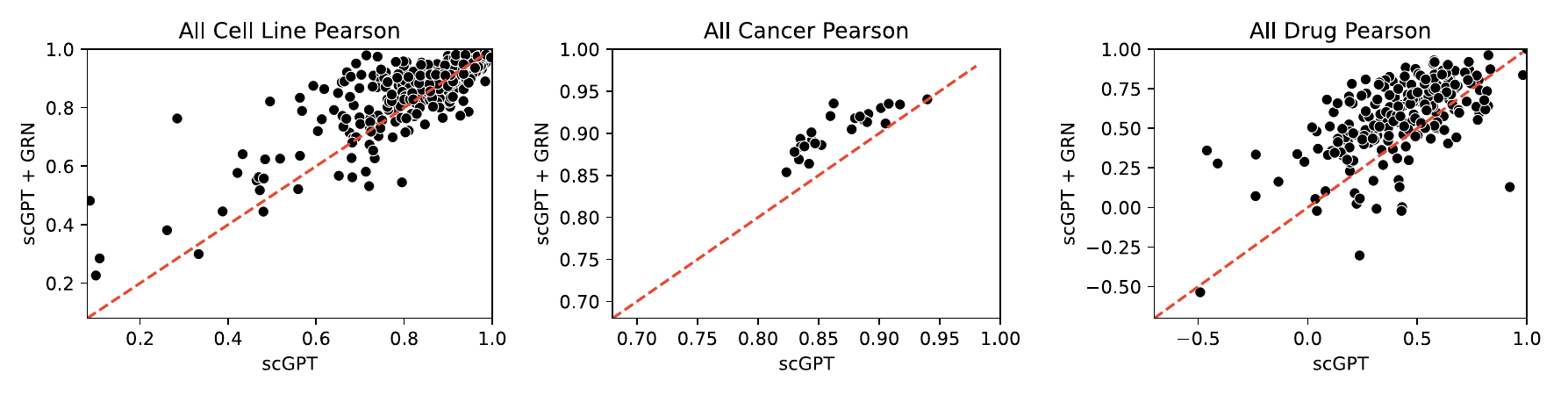}
        \vspace{-10mm}
        \caption{Pairwise visualization of the Pearson correlation coefficient
        of scFoundation and scPaLM based on different grouping strategies. Left:
        grouping with respect to the cell lines; Middle: grouping with respect
        to the cancer type; Right: grouping with respect to the drug type. The red
        lines indicate the relationship of $y = x$.}
        \label{fig:drugresponse}
        \vspace{-4mm}
    \end{figure*}

    Our evaluation utilized data from the Cancer Cell Line Encyclopedia (CCLE)
    and Genomics of Cancer Drug Sensitivity (GDSC) databases~\cite{iorio2016landscape,
    barretina2012cancer}. The model integrates gene expression profiles with drug
    structural information to predict drug sensitivity. As shown in Fig.~\ref{fig:drug_response_num},
    our GRN-enhanced approach achieves superior performance across different
    experimental settings. Our model achieves a correlation coefficient of $0.906
    \pm0.0.002$, significantly outperforming both DeepCDR ($0 .838\pm 0.0 01$)
    and the baseline scGPT model ($0.875\pm 0.010$). Furthermore, as shown in
    Fig.~\ref{fig:drugresponse}, our GRN-integrated model demonstrates superior
    performance over the baseline across all cancer types. The enhanced model exhibits
    consistently better predictive capability than baseline approaches under most
    cell lines and drug conditions, achieving robust performance improvements
    across different experimental settings.

    \subsection{Single-Cell Drug Response Classification}

    Single-cell drug response classification presents a unique challenge in
    cancer research, requiring drug sensitivity prediction at the individual cell
    resolution. This task is particularly challenging due to the limited
    availability of single-cell drug response data and the need to transfer
    knowledge from bulk-level pharmacogenomic data to single cells~\cite{zheng2023enabling,
    hao2023large}.

    \begin{table}[!ht]
        \centering
        \vspace{-2mm}
        \caption{Single-cell drug response classification. Superior model between
        backbone and GRN (ours) is bolded, while the best performance for each
        drug is underlined.}
        \vspace{-3mm}
        \label{tab:sc_drug_response} \resizebox{\linewidth}{!}{\begin{tabular}{lccccc}
            \toprule Model                              & Etoposide           & NVP-TAE684                   & PLX4720                      & Sorafenib                    & Avg. AUC $\uparrow$                                   \\
            \hline
            SCAD                                        & \underline{$0.696$} & $0.613$                      & $0.380$                      & $0.572$                      & $0.565$\scriptsize{$\pm{0.134}$}                      \\
            \midrule scGPT                              & $\textbf{0.511}$    & $0.415$                      & $0.563$                      & $0.346$                      & $0.459$\scriptsize{$\pm{0.097}$}                      \\
            \rowcolor[gray]{0.90} + \textbf{GRN (ours)} & $0.510$             & $\textbf{0.663}$             & $\textbf{0.678}$             & $\textbf{0.474}$             & $\textbf{0.581}$\scriptsize{$\pm{0.104}$}             \\
            \midrule scFoundation                       & $0.596$             & $0.750$                      & $\textbf{0.694}$             & $0.807$                      & $0.712$\scriptsize{$\pm{0.090}$}                      \\
            \rowcolor[gray]{0.90} + \textbf{GRN (ours)} & $\textbf{0.663}$    & $\underline{\textbf{0.760}}$ & $0.598$                      & \underline{$\textbf{0.953}$} & \underline{$\textbf{0.743}$}\scriptsize{$\pm{0.155}$} \\
            \midrule scPaLM                             & $0.471$             & $\textbf{0.730}$             & $0.502$                      & $0.299$                      & $0.500$\scriptsize{$\pm{0.177}$}                      \\
            \rowcolor[gray]{0.90} + \textbf{GRN (ours)} & $\textbf{0.483}$    & $0.468$                      & \underline{$\textbf{0.689}$} & $\textbf{0.602}$             & $\textbf{0.561}$\scriptsize{$\pm{0.105}$}             \\
            \bottomrule
        \end{tabular}}
        \vspace{-2mm}
    \end{table}

    We evaluated our model on four drugs (Sorafenib, NVP-TAE684, PLX4720, and Etoposide).
    Performance was assessed using the Area Under the ROC Curve (AUC) for classification
    accuracy. Table~\ref{tab:sc_drug_response} demonstrates \ourmethod's
    superiority across most settings. Benefiting from the integration of GRN information,
    our model achieved a $4.4\%$ performance improvement on scFoundation, surpassing
    the previous SOTA. For each drug, we report average performance metrics computed
    through five-fold cross-validation.

    \subsection{Ablation Studies}
    \label{sec:ablation}

    \textbf{Effectiveness of GRN Types.} We first investigate how different GRN construction
    strategies influence model performance. We evaluate four variants: (1) \textit{Random
    GRN}: Randomly generated networks with matched edge counts; (2) \textit{Cell-type
    Specific}: GRNs constructed using SCENIC+ at cell population level; (3)
    \textit{Cell-specific}: Single-cell resolution GRNs via AUCell thresholding;
    (4) \textit{Hybrid}: Our proposed combination of cell-type and cell-specific
    GRNs. Experiments are conducted on the scGPT backbone with identical
    hyperparameters across all variants.

    \begin{table}[h]
        \vspace{-2mm}
        \caption{Variants of GRN types {\small (Backbone: scGPT)}}
        \vspace{-3mm}
        \label{tab:grn_types}
        \centering
        \resizebox{0.7\columnwidth}{!}{%
        \begin{tabular}{lcc}
            \toprule GRN Type                                     & Drug Response PCC $\uparrow$             \\
            \midrule No GRN                                       & 0.875 \scriptsize{$\pm{0.010}$}          \\
            Random                                                & 0.892 \scriptsize{$\pm{0.006}$}          \\
            Cell-type Specific                                    & 0.901 \scriptsize{$\pm{0.003}$}          \\
            Cell-specific                                         & 0.902 \scriptsize{$\pm{0.002}$}          \\
            \midrule \rowcolor[gray]{0.90} \textbf{Hybrid (Ours)} & \textbf{0.906} \scriptsize{$\pm{0.002}$} \\
            \bottomrule
        \end{tabular}%
        }
        \vspace{-2mm}
    \end{table}


    Table~\ref{tab:grn_types} demonstrates that our hybrid approach achieves
    superior performance, with relative improvements of $0.5\%$ in drug response
    prediction compared to single-scale GRNs. The cell-specific and cell-type-specific
    variants show better performance than random networks, suggesting the importance
    of capturing regulatory information.

    \noindent
    \textbf{Impact of Edge Perturbation Strategies.} We next analyze the
    effectiveness of our biologically informed edge perturbation strategy. Two variants
    are compared: (1) \textit{Random Perturbation}: $20\%$ edges randomly
    replaced; (2) \textit{Co-expression Guided}: Our proposed strategy using gene
    co-expression patterns. Experiments are conducted on scPaLM using identical
    training protocols.

    \begin{table}[h]
        \vspace{-2mm}
        \caption{Edge perturbations {\small(Backbone: scPaLM)}}
        \vspace{-3mm}
        \label{tab:perturbation}
        \centering
        \resizebox{0.95\columnwidth}{!}{%
        \begin{tabular}{lcc}
            \toprule Edge perturbation strategies                               & Drug Response PCC $\uparrow$             & Response Classification AUC $\uparrow$   \\
            \midrule No Augmentation                                            & 0.870 \scriptsize{$\pm{0.006}$}          & 0.555 \scriptsize{$\pm{0.113}$}          \\
            Random Perturbation                                                 & 0.867 \scriptsize{$\pm{0.002}$}          & 0.548 \scriptsize{$\pm{0.108}$}          \\
            \midrule \rowcolor[gray]{0.90} \textbf{Co-expression Guided (Ours)} & \textbf{0.884} \scriptsize{$\pm{0.004}$} & \textbf{0.561} \scriptsize{$\pm{0.105}$} \\
            \bottomrule
        \end{tabular}%
        }
        \vspace{-2mm}
    \end{table}

    As shown in Table~\ref{tab:perturbation}, our co-expression guided
    perturbation achieves $1.6\%$ relative improvements over the baseline in the
    drug response prediction tasks. It is noteworthy that simple random perturbation-based
    data augmentation may degrade model performance, highlighting the necessity
    of our co-expression guided perturbation strategy.

    \noindent
    \textbf{Analysis of GNN Architectures}. We further examine how different GNN
    architectures affect model performance when integrated with scFoundation. We
    compare three popular GNN variants: (1) \textit{GCN}: Standard graph
    convolutional networks~\cite{kipf2016semi}; (2) \textit{GIN}: Graph isomorphism
    networks~\cite{xu2018powerful}; (3) \textit{GraphSAGE}: Our choice with the
    neighbor sampling approach.

    \begin{table}[!ht]
        \vspace{-2mm}
        \caption{Variants of GNN types {\small (Backbone: scFoundation)}}
        \vspace{-3mm}
        \label{tab:gnn_types}
        \centering
        \resizebox{1\columnwidth}{!}{%
        \begin{tabular}{lccc}
            \toprule GNN Type                                        & Drug Response PCC $\uparrow$            & Response Classification AUC $\uparrow$   \\
            \midrule GCN                                             & 0.881 \scriptsize{$\pm{0.007}$}         & 0.675 \scriptsize{$\pm{0.014}$}          \\
            GIN                                                      & 0.876 \scriptsize{$\pm{0.006}$}         & 0.623 \scriptsize{$\pm{0.138}$}          \\
            \midrule \rowcolor[gray]{0.90} \textbf{GraphSAGE (Ours)} & \textbf{0.888}\scriptsize{$\pm{0.002}$} & \textbf{0.743} \scriptsize{$\pm{0.155}$} \\
            \bottomrule
        \end{tabular}%
        }
        \vspace{-2mm}
    \end{table}

    Table~\ref{tab:gnn_types} reveals that GraphSAGE performs best while maintaining
    computational efficiency. The $1.4\%$ improvement in response prediction
    over GIN demonstrates the effectiveness of neighbor sampling for handling
    GRN sparsity.

    \subsection{Analysis of Attention Patterns}
    \label{sec:attention_analysis}

    To investigate how our model leverages gene regulatory relationships, we
    analyze the attention patterns in the cross-attention fusion layer. Let $\mathbf{A}
    ^{(h)}\in \mathbb{R}^{N \times N}$ denote the attention matrix for head $h$ in
    the multi-head cross-attention mechanism, where $N$ is the number of genes. Each
    entry $a^{(h)}_{ij}$ represents the attention weight between query gene $i$ (from
    the RNA FM) and key gene $j$ (from the GNN encoder). We compute the \textit{gene-wise
    attention importance score} $\phi_{j}$ for each gene $j$ by averaging across
    all heads and query genes as $\phi_{j}= \frac{1}{H \cdot N}\sum_{h=1}^{H}\sum
    _{i=1}^{N}a^{(h)}_{ij}$, where $H$ is the number of attention heads. This score
    quantifies how frequently a gene's regulatory embedding influences other genes'
    expression representations.

\begin{figure}[t]
        \centering
        \includegraphics[width=0.95\columnwidth]{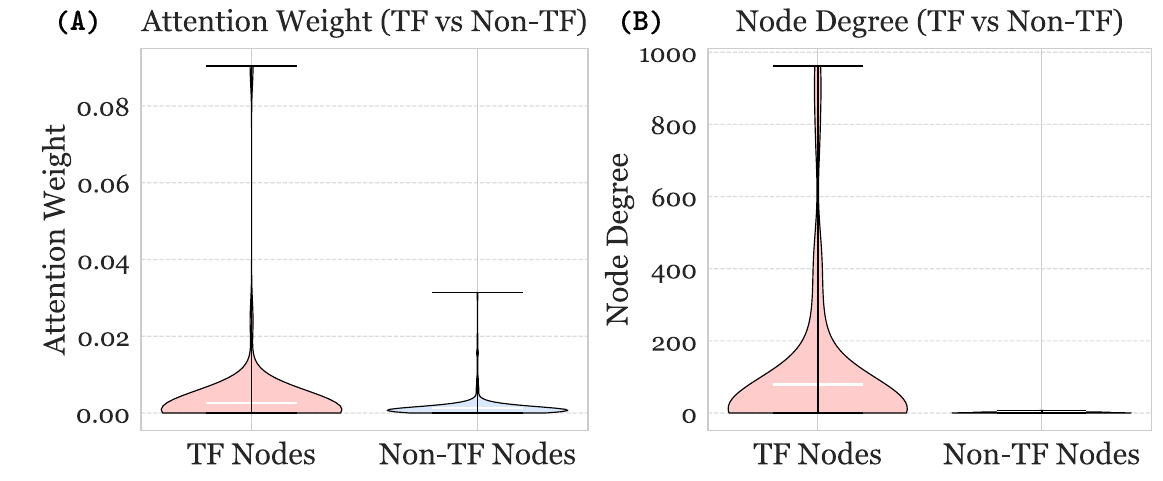}
        \vspace{-2mm}
        \caption{(A) Distribution of average attention scores for transcription
        factor (TF) and non-transcription factor (non-TF) nodes; (B) Node degree
        distributions for these two types of nodes. TF nodes appear to connect
        to more genes and also exhibit higher attention weights.}
        \label{fig:attn_analysis}
        \vspace{-3mm}
    \end{figure}

    To identify biologically meaningful patterns, we calculate the \textit{transcription
    factor (TF) enrichment ratio} $\rho$:
    \begin{equation}
        \rho = \frac{\mathbb{E}[\phi_{j}| j \in \mathcal{T}]}{\mathbb{E}[\phi_{j}|
        j \notin \mathcal{T}]},
    \end{equation}
    where $\mathcal{T}$ denotes the set of transcription factors in our GRNs.
    $\rho > 1$ indicates preferential attention to TFs. Our analysis reveals $\rho
    = 2.011$ across all cell types on the drug response prediction task,
    indicating the model attends disproportionately to TFs. The distributions of
    node degrees for TF and non-TF nodes, as well as the cross attention weights
    in the fusion layer, are shown in Figure \ref{fig:attn_analysis}.

    \section{Conclusion}
    In this paper, we propose \textbf{\ourmethod}, a framework that systematically
    integrates \textit{multi-scale gene regulatory networks} into RNA foundation
    models through two key innovations: (1) hierarchical GRN construction via multi-omics
    fusion, and (2) a structure-aware adapter combining adaptive cross-attention
    with biologically informed edge perturbation to resolve the topological
    imbalance. \ourmethod achieves consistent performance improvement across therapeutic
    development tasks. Attention analysis reveals biologically meaningful
    patterns of our edge perturbation strategy. The framework's universal applicability
    is validated through the integration with major RNA foundation architectures,
    establishing a new paradigm for biologically grounded AI in computational
    genomics.

    \section*{Limitations}

    \noindent
    \textbf{Dependency on Regulatory Databases.} The quality of our constructed
    GRN relies heavily on existing motif databases and chromatin accessibility
    data. Similar to SCENIC+~\cite{bravo2023scenic+}, our approach cannot fully resolve
    ambiguous TF binding patterns within shared motif families. Future
    integration of emerging techniques like GET-style pseudobulk chromatin
    profiles~\cite{fu2025foundation} probably could further improve the
    reliability of gene regulatory information.

    \noindent
    \textbf{Multi-modal Data Requirement.} While our framework theoretically supports
    single-modality data through reference mapping, optimal GRN construction
    requires paired scRNA-seq/scATAC-seq data. Future work could try integrate lifelong
    learning strategies to reduce multi-modal dependency through atlas-scale data
    integration~\cite{yuan2024inferring}. Additionally, inspired by GET~\cite{fu2025foundation},
    constructing pseudo-paired multi-omics data from existing resources may
    better leverage heterogeneous datasets.

    \section*{Ethics Statement}
    Our work on integrating multi-scale gene regulatory networks into RNA
    foundation models demonstrates a commitment to advancing biomedical AI while
    adhering to ethical research practices. All datasets used in this study listed
    in Table~\ref{tab:dataset_summary} are publicly available and fully anonymized,
    with all donor identities and sensitive metadata removed in compliance with
    privacy regulations. While our model shows promise in accelerating drug discovery
    and improving gene therapies, any clinical application must undergo rigorous
    ethical review to ensure compliance with genomic data protection standards. We
    emphasize that biological foundation models built upon our methodology should
    incorporate safeguards against misuse, such as restricting access to
    potentially harmful gene-editing predictions. Furthermore, our
    implementation prioritizes transparency—all code and preprocessing workflows
    are designed for public auditability, reproducibility, and explainability.

    \section*{Acknowledgment}
    This manuscript has been authored by Lawrence Livermore National Security, LLC
    under Contract No. DE-AC52-07NA27344 with the U.S. Department of Energy. The
    United States Government retains, and the publisher, by accepting the
    article for publication, acknowledges that the United States Government retains
    a non-exclusive, paid-up, irrevocable, worldwide license to publish or
    reproduce the published form of this manuscript, or allow others to do so,
    for United States Government purposes.

    This research was, in part, funded by the National Institutes of Health (NIH)
    under other transactions 1OT2OD038045-01. The views and conclusions
    contained in this document are those of the authors and should not be
    interpreted as representing official policies, either expressed or implied,
    of the NIH.

    \bibliography{acl_latex}

    \onecolumn
    \appendix

    \section{Cell-type-specific GRNs via eRegulon Inference.}
    \label{appendix:eregulon} We construct hierarchical GRNs using SCENIC+~\cite{bravo2023scenic+},
    which integrates scATAC-seq and scRNA-seq through three phases:

    \ding{182} \textit{Candidate Enhancer Identification:} Chromatin
    accessibility profiles from scATAC-seq reveal genomic regions where DNA is
    unwound, indicating potential regulatory elements. \textbf{Co-accessible
    regions} are detected using pycisTopic~\cite{bravo2019cistopic}, which
    employs topic modeling – a probabilistic method that groups genomic loci
    with similar accessibility patterns across cells. These regions, enriched
    near genes with correlated expression, serve as candidate enhancers – non-coding
    DNA elements that promote gene transcription.

    \ding{183} \textit{TF-Motif Enrichment Analysis:} Transcription factors bind
    DNA through specific sequence patterns called \textbf{motifs} (e.g., the E-box
    "CANNTG" for basic helix-loop-helix TFs~\cite{wright1992muscle,malik1995role}).
    Enhancer candidates are scanned against a curated database of $32,765$ TF-binding
    motifs (aggregated from $29$ collections~\cite{bravo2023scenic+}) using pycisTarget.
    Two algorithms identify statistically overrepresented motifs: $i$) The
    \textit{cisTarget} algorithm ranks motifs by how early their target regions appear
    in accessibility-based rankings; $ii$) The \textit{DEM} algorithm identifies
    motifs differentially enriched between cell types. These algorithms establish
    \textbf{TF-to-enhancer links} (NES $>3.0$, FDR $<0.1$) 
    while mitigating false positives through motif clustering.

    \ding{184} \textit{eRegulon Construction:} For each TF, we link its target enhancers
    to genes using three criteria: (1) genomic proximity ($\pm 150$kb from gene),
    (2) expression correlation (Pearson $|r|>0.03$), and (3) gradient-boosted regression
    importance scores (GRNBoost2~\cite{moerman2019grnboost2}). This forms
    \textbf{enhancer-driven regulons (eRegulons)} – triplets connecting TFs,
    enhancers, and target genes that function as regulatory units. Cell-type
    specificity is determined by joint accessibility of enhancers and expression
    of target genes~\cite{bravo2023scenic+}.

    \section{Algorithm}
    \label{appendix:alg}
    Algorithm~\ref{alg:pretrain} formalizes our structure-aware pretraining
    process, implementing the key components described in \S\ref{sec:structure_adapter}
    and \S\ref{sec:inference_pipeline}. The pseudocode explicitly shows the edge
    perturbation strategy (Lines $3-12$) that addresses topological imbalance
    through co-expression guided augmentation, and the multi-scale fusion mechanism
    (Lines $14-20$) combining cell-specific and cell-type-specific GRN embeddings.
    This algorithm complements Fig.~\ref{fig:overview} in the main text by detailing
    how biological priors are injected during training while maintaining compatibility
    with various backbone architectures.

    \begin{algorithm}
        [H] \small
        \caption{\small{Structure-Aware (Continue) pretraining with Multi-scale GRN}}
        \label{alg:pretrain}
        \begin{algorithmic}
            [1] \State \textbf{Input:} Masked gene expression vector $x$, cell-specific
            GRN $G_{\text{cell}}$, cell-type-specific GRN $G_{\text{type}}$, GNN
            encoder $F$, Transformer backbone $H$, cross-attention module $P$,
            perturbation ratio $\alpha$, fusion weight $\beta$ \State \textbf{Output:}
            Reconstructed expression $\bar{x}$

            \Function{PerturbGRN}{$G, G_{\text{co}}, \alpha$} \State $V \gets \text{nodes}
            (G)$ \State $E_{\text{original}}\gets \text{edges}(G)$ \State $E_{\text{drop}}
            \gets \text{Sample}(E_{\text{original}}, \alpha|E_{\text{original}}|)$
            \State $E_{\text{co}}\gets \text{Sample}(\text{edges}(G_{\text{co}}),
            \alpha |E_{\text{original}}|)$ 
            \State \Return
            $(V, E_{\text{original}}\setminus E_{\text{drop}}\cup E_{\text{co}})$
            \EndFunction

            \State \textit{// Stage 1: Graph Augmentation} \State
            $G_{\text{co}}\gets \text{ConstructCoExpressionGraph}(x)$ 
            \State $\tilde{G}_{\text{cell}}\gets \Call{PerturbGRN}{G_{\text{cell}}, G_{\text{co}}, \alpha}$
            \State $\tilde{G}_{\text{type}}\gets \Call{PerturbGRN}{G_{\text{type}}, G_{\text{co}}, \alpha}$

            \State \textit{// Stage 2: Structural Encoding} \State $h_{\text{cell}}
            \gets F(\tilde{G}_{\text{cell}})$ \Comment{Cell-specific encoding}
            \State $h_{\text{type}}\gets F(\tilde{G}_{\text{type}})$ \Comment{cell-type-specific encoding}
            \State $h_{\text{struct}}\gets h_{\text{cell}}\oplus h_{\text{type}}$
            \Comment{Element-wise sum}

            \State \textit{// Stage 3: Cross-modal Fusion} \State
            $h_{\text{expr}}\gets H(x)$ \Comment{Gene expression embedding} \State
            $h_{\text{fusion}}\gets P(h_{\text{expr}}, h_{\text{struct}})$ \Comment{Cross-attention fusion}
            \State
            $h_{\text{combined}}\gets h_{\text{expr}}+\beta h_{\text{fusion}}$
            \Comment{Weighted combination}

            \State $\bar{x}\gets \text{Decoder}(h_{\text{combined}})$ \State
            \Return $\bar{x}$
        \end{algorithmic}
    \end{algorithm}
    \clearpage
    \section{Datasets}
    \label{appendix:datasets} Table~\ref{tab:dataset_summary} summarizes key statistics
    for all experimental datasets. The SEA-AD multiome dataset provides paired
    scRNA-seq/scATAC-seq profiles for pretraining, while the perturbation benchmarks
    (Adamson, Dixit, Norman) and drug response datasets (CCLE, GDSC) enable
    comprehensive downstream evaluation across different biological contexts.
    \begin{table*}
        [!hbtp]
        \centering
        \vspace{-2mm}
        \caption{\small Summary of datasets used in different tasks.}
        \resizebox{0.95\linewidth}{!}{
        \begin{tabular}{c|c|c|c}
            \toprule Task                                          & Dataset                                       & \# of cells/\# of cell lines & \# of genes    \\
            \midrule Training: Mask Language Modeling              & SEA-AD (multiome part)\cite{hawrylycz2024sea} & $113,209$                    & $18,984$       \\
            \midrule \multirow{3}{*}{Gene Perturbation Prediction} & Adamson\cite{adamson2016multiplexed}          & $68,603$                     & $5,060$        \\
                                                                   & Dixit\cite{dixit2016perturb}                  & $447,35$                     & $5,012$        \\
                                                                   & Norman\cite{norman2019exploring}              & $91,205$                     & $5,045$        \\
            \midrule Drug Response Prediction/                     & CCLE\cite{barretina2012cancer}                & $947$                        & $1651$         \\
            Single Cell Drug Response Classification               & GDSC\cite{iorio2016landscape}                 & $969$                        & $\sim{}22,000$ \\
            \bottomrule
        \end{tabular}
        }
        \label{tab:dataset_summary}
        \vspace{-4mm}
    \end{table*}


    \section{Transcription Factor Activity Distribution}
    \label{appendix:tf_dist} 
\begin{figure*}[!h]
        \centering
        \includegraphics[width=\textwidth]{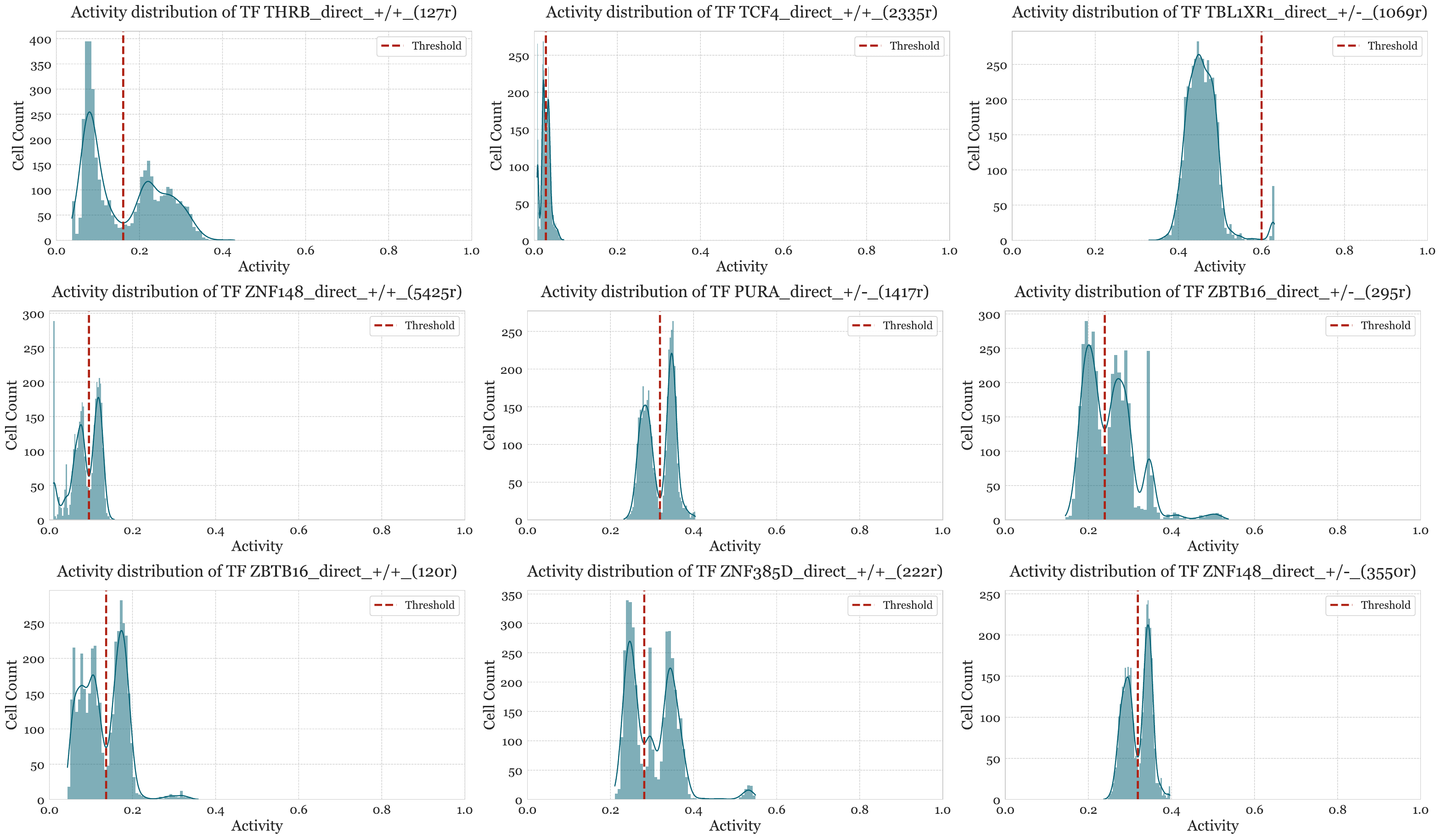}
        \vspace{-8mm}
        \caption{The distribution of activity levels for nine randomly selected
        transcription factors (TFs) within a single cell type. The threshold distinguishing
        active versus inactive states are demarcated by red vertical lines.}
        \label{fig:aucell}
        \vspace{-4mm}
    \end{figure*}

Fig.~\ref{fig:aucell} visualizes the bimodal and
    skewed AUC distributions underlying the single-cell GRN construction, supporting
    the thresholding methodology from \S\ref{sec:grn_construction}. The clear separation
    of active/inactive states for TFs like PURA empirically validates the gaussian
    mixture modeling approach. These distributions directly inform the cell-specific
    regulatory networks that drive our model's performance improvements in
    downstream tasks (\S\ref{sec:ablation}).

    \clearpage
    \section{Potential Risks}
    \label{app:risk} While \ourmethod advances computational genomics, three key
    risks warrant consideration: \textbf{(1) Data Bias Propagation:} Reliance on
    existing motif databases may propagate biases in TF-gene interactions, particularly
    for understudied cell types or minor populations, potentially leading to
    skewed therapeutic predictions. \textbf{(2) Privacy Vulnerabilities:}
    Although using anonymized data, integration of multi-omics profiles could theoretically
    enable cell identity re-identification through rare regulatory signatures.
    \textbf{(3) Dual-Use Concerns:} Enhanced prediction of gene regulatory outcomes
    might be misused to design targeted biological agents, though our current
    implementation focuses only on therapeutic contexts. We mitigate these risks
    through (1) transparent documentation of data sources, and (2) controlled
    access to regulatory network components. Responsible deployment requires ongoing
    collaboration with bioethicists and clinical reviewers.

\end{document}